# Concept-Oriented Deep Learning: Generative Concept Representations

Daniel T. Chang (张遵)

*IBM (Retired)* dtchang43@gmail.com

**Abstract:** Generative concept representations have three major advantages over discriminative ones: they can represent uncertainty, they support integration of learning and reasoning, and they are good for unsupervised and semi-supervised learning. We discuss probabilistic and generative deep learning, which generative concept representations are based on, and the use of variational autoencoders and generative adversarial networks for learning generative concept representations, particularly for concepts whose data are sequences, structured data or graphs.

## 1 Introduction

Concept-oriented deep learning (CODL) [1] extends deep learning with concept representations and conceptual understanding capability. The concept representations discussed in [1] are deterministic and discriminative in nature. In this paper we discuss generative concept representations, which are probabilistic and generative in nature, for CODL. Generative concept representations have three major advantages: they can represent uncertainty, they support integration of deep learning and reasoning, and they are good for unsupervised and semi-supervised learning.

### 1.1 Uncertainty

*Uncertainty* [2] arises because of limitations in our ability to observe the world, limitations in our ability to model it, and possibly inherent non-determinism (e.g., quantum phenomena). From the perspective of deep learning, there are two types of uncertainty: model uncertainty and data uncertainty [3, 4]. Model uncertainty accounts for uncertainty in the model structure and model parameters. This is important to consider for safety-critical applications and small datasets. Data uncertainty accounts for out-of-distribution data and noisy data. This is important to consider for real-time applications and large datasets.

Uncertainty, therefore, is an inescapable aspect of most real-world data-driven applications [2]. Because of this, we need to allow our learning system to represent uncertainty and our reasoning system to consider different possibilities due to uncertainty. Further, to obtain meaningful conclusions, we need to reason not just about what is possible, but also about what is probable. *Probability theory* provides us with a formal framework for representing uncertainty and for considering multiple possible outcomes and their likelihood.

## 1.2 Integration of Deep Learning and Reasoning

*Deep learning* has achieved significant success in many perception and learning tasks. However, a real AI system must additionally possess the ability of *reasoning* to be useful. To achieve integrated intelligence that involves perception, learning and reasoning, it is desirable to tightly integrate deep learning (for learning and perception tasks) and *probabilistic graphical models* (for reasoning tasks) within a principled probabilistic framework [5].

To ensure efficient and effective information exchange between the learning and perception component and the reasoning component, a natural way is to represent the learning and perception component as a probabilistic graphical model and seamlessly connect it to the reasoning probabilistic graphical model. Such integration has been successfully used for molecular design [6].

## 1.3 Unsupervised and Semi-supervised Learning

A shortcoming of the current state of the art of supervised deep learning is that it requires large amounts of labeled data to achieve good accuracy. Reducing the amount of labeled data necessary for deep learning to work well and be applicable across a broad range of tasks requires *unsupervised learning* or *semi-supervised learning*. However, a central cause of the difficulties with unsupervised and semi-supervised learning is the high dimensionality of the random variables to be considered. This brings two distinct challenges [7]: a statistical challenge and a computational challenge. The statistical challenge regards generalization; the computational challenge arises because many algorithms involve intractable computations.

*Deep generative models* have been proposed to avoid these intractable computations by design [7].

## 2 Probabilistic and Generative Deep Learning

Whereas concept representations are based on traditional deep learning, which is deterministic and discriminative in nature, generative concept representations are based instead on *probabilistic and generative deep learning*, which is discussed in this section.

### 2.1 Probabilistic Machine Learning

The key idea behind *probabilistic machine learning* [8] is that learning can be thought of as inferring probabilistic models to explain observed data. A machine can use such models to make predictions about future data, and take actions that are rational given these predictions. Uncertainty plays a fundamental part in all of this and probability theory provides a framework for modeling uncertainty, as mentioned previously.

The probabilistic approach to modeling uses probability theory to represent all forms of uncertainty:

- *Probability distributions* are used to represent all uncertain elements in a model (including structural, parametric, out-of-distribution and noise-related) and how they relate to the data.

- The basic rules of probability theory are used to infer the uncertain elements given the observed data.

- Learning from data occurs through the transformation of the *prior distributions* (defined before observing the data) into *posterior distributions* (after observing the data). The application of probability theory to learning from data is called *Bayesian learning* [8, 9].

Simple probability distributions over single or a few *random variables* can be composed to form the building blocks of larger, more complex models. The compositionality of probabilistic models means that the behavior of these building blocks in the context of the larger model is often much easier to understand. The dominant paradigm for representing such compositional probabilistic models is *probabilistic graphical models*, which will be discussed in the next subsection.

There are two main challenges [8] for probabilistic machine learning. The main modeling challenge is that the model should be flexible enough to capture all the properties of the data required to achieve the prediction task of interest. (The key statistical concept underlying flexible models that grow in complexity with the data is non-parametrics [9].) Computationally, the main challenge is that learning involves marginalizing (summing out) all the variables in the model except for the variables of interest. Such high-dimensional sums and integrals are generally computationally hard.

## 2.2 Probabilistic Graphical Models

*Probabilistic graphical models (PGMs)* [2, 7, 10] are structured probabilistic models. A PGM is a way of describing a probability distribution of random variables, using a *graph* to describe which variables in the probability distribution directly interact with each other, with each *node* represents a variable and each *edge* represents a direct interaction. This allows the models to have significantly fewer parameters which can in turn be estimated reliably from less data. These smaller models also have dramatically reduced computational cost in terms of storing the model, performing inference in the model, and drawing samples from the model.

Usually a PGM comes with both a *graphical representation* of the model and a *generative process* to depict how the random variables are generated. Due to its Bayesian nature, PGM is easy to extend to incorporate other information or to perform other tasks.

*Directed PGMs*

There are essentially two types of PGM [2]: directed PGM (also known as Bayesian networks) and undirected PGM (also known as Markov random fields). *Directed PGMs* are based on directed graphs (the direction of the arrow indicating which variable's probability distribution is defined in terms of the other's) and are useful because both the structure and the parameters provide a natural representation for many types of real-world domains. Undirected PGMs, on the other hand, are useful in modeling a variety of phenomena where one cannot naturally ascribe directionality to the interaction between variables. We focus on directed PGMs.

A directed PGM defined on *random variables* $x$ is defined [7] by a directed acyclic graph G whose vertices are the random variables in the model, and a set of local conditional probability distributions $p(x_i | Pa_G(x_i))$ where $Pa_G(x_i)$ gives the parents of $x_i$ in G. The *probability distribution over $x$* is given by

$$p(\mathbf{x}) = \Pi_i \, p(x_i | Pa_G(x_i)).$$

So long as each variable has few parents in the graph, the distribution can be represented with very few parameters. Some restrictions on the graph structure, such as requiring it to be a tree, can also guarantee that operations like computing marginal or conditional distributions over subsets of variables are efficient. The directed PGM syntax does not place any constraint on how we define our conditional distributions. It only defines which variables they are allowed to take in as arguments.

*Latent Variables*

A good PGM needs to accurately capture the distribution over the *observed variables $x$*. Often the different elements of $\mathbf{x}$ are highly dependent on each other. The approach most commonly used to model these dependencies is to introduce several *latent variables* $\mathbf{z}$ [7]. The model can then capture dependencies between any pair of variables $x_i$ and $x_j$ indirectly, via direct dependencies between $x_i$ and $\mathbf{z}$, and direct dependencies between $\mathbf{z}$ and $x_j$.

Latent variables have advantages beyond their role in efficiently capturing p(**x**). The latent variables **z** also provide an alternative representation for **x**. Many approaches accomplish *representation learning* by learning latent variables.

*The Deep Learning Approach*

The *deep learning approach* [7] tends to use different model structures, learning algorithms and inference procedures than are commonly used by traditional probabilistic graphical modeling. Deep learning PGMs typically have more latent variables than observed variables. Complicated nonlinear interactions between observed variables are accomplished via indirect connections that flow through multiple latent variables. Also, the deep learning approach does not intend for the latent variables to take on any specific semantics ahead of time—the training algorithm is free to invent the "concepts" it needs to model a particular dataset. The latent variables are usually not very easy to interpret.

In the context of PGM, we can define the depth of a model in terms of the *graphical model depth* rather than the computational graph depth. We can think of a latent variable $z_i$ as being at depth j if the shortest path from $z_i$ to an observed variable is j steps. We usually describe the depth of the model as being the greatest depth of any such $z_i$. We focus on *deep PGM*s, which are PGMs with multiple layers of latent variables.

## 2.3 Deep Generative Models

*Unsupervised and semi-supervised learning* methods are often based on *generative models* [11-13], which are probabilistic models that express hypotheses about the way in which data may have been generated. *PGMs* have emerged as a broadly useful approach to specifying generative models. The elegant marriage of graph theory and probability theory makes it possible to take a fully probabilistic approach to unsupervised and semi-supervised learning in which efficient algorithms are available to update a *prior generative model* into a *posterior generative model* once data have been observed.

All of the generative models [7] represent probability distributions over multiple variables in some way. Some allow the probability distribution function to be evaluated explicitly. Others do not allow the evaluation of the probability distribution function, but support operations that require knowledge of it, such as drawing samples from the distribution. Some of these models are PGMs. We focus on *deep generative models (DGMs)*, which are *deep PGMs* and/or employ *deep neural networks* for parameterizing the models.

*Prescribed and Implicit DGMs*

As mentioned above, there are two types of DGMs [12]: prescribed DGMs and implicit DGMs. *Prescribed DGMs* are those that provide an explicit parametric specification of the probability distribution of the observed variable **x**, specifying the *likelihood function $p_\theta(x)$* with parameter **θ**. *Implicit DGMs* are defined by a stochastic procedure (a simulation process) that generates data - we can sample data from its generative process, but we may not have access to calculate its probability distribution, and thus are referred to as *likelihood-free*.

## 2.4 Differentiable Generator Networks

Many *DGMs* are based on the idea of using a *differentiable generator network* [7]. The model transforms samples of latent variables **z** to observed variables **x** or to distributions over **x** using a *differentiable function* $g(z; \theta^{(g)})$ which is typically represented by a *neural network*. This model class includes

- *variational autoencoders* which pair the differentiable generator network with an *inference network*, and

- *generative adversarial networks* which pair the differentiable generator network with a *discriminator network*.

In each case, the paired networks are jointly trained.

Generator networks are essentially parameterized computational procedures for generating samples, where the *architecture* provides the family of possible distributions to sample from and the *parameters* select a distribution from within that family. To generate samples from complicated distributions that are difficult to specify directly, difficult to integrate over, or whose resulting integrals are difficult to invert, one uses a *neural network* to represent a parametric family of nonlinear functions $g(z; \theta^{(g)})$, and uses training data to infer the parameters selecting the desired function.

Approaches based on differentiable generator networks are motivated by the success of *gradient descent* applied to differentiable neural networks for classification and regression. However, generative modeling is more difficult than classification or regression because the learning process requires optimizing intractable criteria. In the case of generative modeling, the learning procedure needs to determine how to arrange **z** space in a useful way and additionally how to map from **z** to **x**.

### 2.5 Generative Concept Representations

*We use DGMs for generative concept representations.* Generative concept representations are *PGMs* and, therefore, can represent uncertainty as well as support integration of deep learning and reasoning. Further, generative concept representations are *generative models* and, therefore, are good for unsupervised and semi-supervised learning.

In the next two sections, we discuss learning generative concept representations using variational autoencoders and generative adversarial networks, two widely-used architectures for DGMs, particularly for concepts whose data are sequences, structured data or graphs.

## 3 Variational Autoencoders for Generative Concept Representations

### 3.1 Autoencoders

An *autoencoder* [7] is a *neural network* that is trained to copy its input to its output. The network may be viewed as consisting of two parts: an *encoder* function $\mathbf{h} = f(\mathbf{x})$ and a *decoder* that produces a reconstruction $\mathbf{x} = g(\mathbf{h})$. Usually they are restricted in ways that allow them to copy only approximately. An autoencoder whose code dimension is less than the input dimension is called undercomplete. Learning an *undercomplete representation* forces the autoencoder to capture the most salient features of the training data.

Modern autoencoders have generalized to stochastic mappings $p_{encoder}(\mathbf{h} \mid \mathbf{x})$ and $p_{decoder}(\mathbf{x} \mid \mathbf{h})$. Theoretical connections between autoencoders and *latent variable models* have brought autoencoders to the forefront of *generative modeling*, as discussed next.

### 3.2 Variational Autoencoders

The *variational autoencoder (VAE)* [7, 11-16] consists of an *encoder network* and a *decoder network* which encodes a data example to a *latent representation* and generates samples from the latent space, respectively. The decoder network is a differentiable generator network and the encoder network is an auxiliary inference network. Both networks are jointly trained using variational learning. *Variational learning* is mainly applied to *prescribed DGMs* and uses the *variational lower bound* of the marginal log-likelihood as the single objective function to optimize both the generator network and the auxiliary inference network.

The likelihood function of a prescribed DGM can be written as [12]:

$$p_\theta(\mathbf{x}) \equiv p_\theta(\mathbf{z})\, p_\theta(\mathbf{x} \mid \mathbf{z})$$

It is usually intractable to directly evaluate and maximize the marginal log-likelihood $\log(p_\theta(\mathbf{x}))$. Following the variational inference approach, one introduces an *auxiliary inference model* $q_\varphi(z \mid x)$ with parameters $\varphi$, which serves as an approximation to the exact posterior $p_\theta(\mathbf{z} \mid \mathbf{x})$.

The variational lower bound can then be derived using the Jenson Inequality:

$$\log(p_\theta(\mathbf{x})) = \log(\sum_z p_\theta(\mathbf{x},\mathbf{z})) = \log(\sum_z q_\varphi(\mathbf{z} \mid \mathbf{x}) \frac{p_\theta(\mathbf{x},\mathbf{z})}{q_\varphi(\mathbf{z} \mid \mathbf{x})}) \geq \sum_z q_\varphi(\mathbf{z} \mid \mathbf{x})\, \log(\frac{p_\theta(\mathbf{x},\mathbf{z})}{q_\varphi(\mathbf{z} \mid \mathbf{x})}) \equiv L(\mathbf{x}; \theta, \varphi)$$

It can further be rewritten as [7, 12]:

$$L(\mathbf{x}; \theta, \varphi) = \sum_z q_\varphi(\mathbf{z} \mid \mathbf{x}) \log(p_\theta(\mathbf{x} \mid \mathbf{z})) - D_{KL}(q_\varphi(\mathbf{z} \mid \mathbf{x}) \,\|\, p_\theta(\mathbf{z}))$$

The first term is the expected reconstruction quality, requiring that $p_\theta(\mathbf{x} \mid \mathbf{z})$ is high for samples of $\mathbf{z}$ from $q_\varphi(\mathbf{z} \mid \mathbf{x})$, while the second term (the KL divergence between the approximate posterior and the prior) acts as a regularizer, ensuring we can generate realistic data by sampling latent variables from $p_\theta(\mathbf{z})$.

Variational learning is to maximize the variational lower bound over the training data. It performs something like the autoencoder, with $q_\varphi(z \mid x)$ *as the encoder* and $p_\theta(x \mid z)$ *as the decoder*. As such, the VAE introduces the constraint on the autoencoder that the latent variable $\mathbf{z}$ is distributed according to a *prior p(z)*. The encoder $q_\varphi(\mathbf{z} \mid \mathbf{x})$ approximates the *posterior p(z | x)*, and the decoder $p_\theta(\mathbf{x} \mid \mathbf{z})$ parameterizes the *likelihood p(x | z)*. The *generation model* is then $\mathbf{z} \sim p(\mathbf{z});\; \mathbf{x} \sim p(\mathbf{x} \mid \mathbf{z})$.

The VAE framework is very straightforward to extend to a wide range of model architectures. This is a key advantage. VAEs also work very well with a diverse family of differentiable operators.

*Concept representations* (latent representations) learned using VAE generally are explicit, continuous and with meaningful structure. As such, they can be directly manipulated to generate new concepts with desired attributes. An example is given in [11] in which *concept vectors* are used to edit images. Further examples will be mentioned later.

## 3.3 Sequence VAEs

*Sequence* data include text (understood as sequences of words), timeseries, molecular structures (represented as SMILES strings), and music (transcribed in MIDI files). They are generally learned using recurrent neural networks. A *recurrent neural network (RNN)* [11, 19] processes sequences by iterating through the sequence elements and maintaining a *state* containing information relative to what it has seen so far. The state of the RNN is reset between processing two different, independent sequences.

Learning with RNNs can be especially challenging due to the difficulty of learning long-term dependencies. The problems of *vanishing and exploding gradients* occur when back-propagating errors across many time steps. To resolve the problems, the *Long-Short Term Memory (LSTM)* architecture [19] introduces the *memory cell*, a unit of computation that replaces traditional nodes in the hidden layer of an RNN. Each memory cell contains a node with a self-connected recurrent edge of fixed weight one, ensuring that the gradient can pass across many time steps without vanishing or exploding.

The *CVAE* [6, 20-21] adapts the VAE to sequence by using single-layer LSTM RNNs for both the encoder and the decoder. *Concept representations* learned using the CVAE contain *latent representation*s of entire sequences and can be used to generate coherent new sequences that interpolate between known sequences. Example usage can be found for text [20], molecular structures [6] and music [21].

There are two main *issues* with the CVAE. First, the decoder, being an RNN, is itself sufficiently powerful to produce an effective model of the data, and therefore the decoder can completely ignore the latent variables. Second, the model must compress the entire sequence to a single latent vector. This begins to fall apart as the sequence length increases. These issues can be overcome by using a *hierarchical RNN for the decoder*, as done in *MusicVAE* [21].

## 3.4 Grammar VAEs

*Structured data* such as symbolic expressions, computer programs, and molecular structures (represented as SMILES strings) have syntax and semantic formalisms. For such data, sequence VAEs will often lead to invalid outputs because of the lack of formalization of syntax and semantics serving as the constraints of the structured data.

*Context-free grammars* can be used to incorporate syntax constraints in the generative model. To do so, structured data are represented as *parse trees* from a context-free grammar. The *grammar variational autoencoder (GVAE)* [24] encodes and decodes directly from and to these parse trees, respectively, ensuring the generated outputs are always valid based on the grammar. It also learns a more coherent latent space in which nearby points decode to similar outputs. GVAE has been applied to symbolic regression and molecular synthesis. Although the context-free grammar provides a mechanism for generating *syntactically valid outputs*, it is incapable to constraint the model for generating semantically valid outputs.

*Attribute grammars* or *syntax-directed definition*s allow one to attach semantics to a parse tree generated by context-free grammar. The *syntax-direct variational autoencoder (SD-VAE)* [25] incorporates attribute grammar in the VAE such that it addresses both syntactic and semantic constraints and generates outputs that are *syntactically valid* and *semantically reasonable*. SD-VAE has been applied to reconstruction and optimization of computer programs and molecules.

### 3.5 Graph VAEs

*Graph* data include knowledge graphs (relations between entities), social networks (relations between people), physical networks (relations between objects), and molecular graphs (representing molecular structures). The *Variational Graph Autoencoder (VGAE)* [27] is a framework for learning graph data. The encoder is a *graph convolutional network* and the decoder is an *inner product decoder*. It makes use of latent variables and is capable of learning interpretable *latent representations* for undirected graphs.

The *Adversarially Regularized Variational Graph Autoencoder (ARVGA)* [28] is an adversarial embedding framework for graph data. The framework extends the above-mentioned *graph convolutional network* to encode the topological structure and node content in a graph to a *latent representation*, on which a *decoder* is trained to reconstruct the graph structure. Furthermore, the latent representation is enforced to match a *prior distribution* via an *adversarial training scheme*. The variational graph encoder learning and adversarial regularization are jointly optimized in a unified framework so that each can be beneficial to the other and lead to a better graph embedding.

## 4 Generative Adversarial Networks for Generative Concept Representations

The *generative adversarial network (GAN)* [7, 11-13, 17-18] consists of a differentiable *generator network* and an auxiliary *discriminator network* which generates samples from *latent representations* and discriminates between data samples and generated samples, respectively. Both networks are jointly trained using adversarial learning. *Adversarial learning* is mainly applied to *implicit DGMs* and optimizes the generator network so that the generated samples are statistically almost indistinguishable from the data samples.

Adversarial learning is based on a game theoretic scenario in which the generator network competes against an adversary [7]:

- The generator network produces samples $\mathbf{x} = g(\mathbf{z}; \boldsymbol{\theta}^{(g)})$.

- Its adversary, the discriminator network, attempts to distinguish between samples drawn from the training data and samples generated by the generator network. The discriminator emits a probability value given by $d(\mathbf{x}; \boldsymbol{\theta}^{(d)})$, indicating the probability that $\mathbf{x}$ is a real data sample rather than a generated sample.

The main motivation for the design of GANs is that the learning process requires neither approximate inference nor approximation of a partition function gradient. Unfortunately, learning in GANs can be difficult in practice. Also, stabilization of GAN learning remains an open problem.

*Concept representations* (latent representations) learned using GAN generally are implicit, discrete and without meaningful structure. As such, they may not be suited for certain applications.

## 4.1 Sequence GANs

Applying GAN to *sequence data* (see 3.3 Sequence VAEs) has *two problems*. First, the generated data from the generator network is based on discrete tokens. The "slight change" guidance from the discriminator network makes little sense because there is probably no corresponding token for such slight change. Second, GAN can only give the score for an entire sequence when it has been generated; for a partially generated sequence, it is non-trivial to balance its current score and the future one once the entire sequence has been generated.

The *sequence generation framework (SeqGAN)* [22] solves the problems by treating the *generator network* as an agent of *reinforcement learning*; the state is the generated tokens so far and the action is the next token to be generated. To give the reward, it employs the *discriminator network* to evaluate the sequence and feedback the evaluation to guide the learning of the generator network. It models the generator network as a *stochastic parameterized policy* and directly trains the policy via policy gradient, which avoids the differentiation difficulty for discrete data in a conventional GAN. SeqGAN has been applied to poem, speech language and music generation.

A common problem in reinforcement learning is *sparse reward*, that the non-zero reward is only observed at the last time step. To deal with sparse reward, the SeqGAN is trained with a stepwise evaluation method, *Monte Carlo tree search*, which stabilizes the training but is computationally intractable when dealing with large dataset. To improve the computational costs, the *stepwise GAN (StepGAN)* [23] uses an alternative stepwise evaluation method to replace Monte Carlo tree search, which automatically assign scores quantifying the goodness of each subsequence at every generation step.

## 4.2 Grammar GANs

For *structured data* (see 3.4 Grammar VAEs), as is the case of sequence VAEs, sequence GANs will often lead to invalid outputs because of the lack of formalization of syntax and semantics serving as the constraints of the structured data. *Context-free grammars* can be used to incorporate syntax constraints in the generative model. The *TreeGAN* [26] incorporate a given context-free grammar into the sequence generation process. In TreeGAN, the generator network employs an RNN to construct a parse tree. The discriminator network uses a tree-structured RNN to distinguish the generated parse trees from sample parse trees. Although the context-free grammar provides a mechanism for generating *syntactically valid outputs*, as is the case of GVAE, it is incapable to constraint the model for generating semantically valid outputs.

In principle, *attribute grammars*, instead of context-free grammars, could be incorporated in the GAN to generate outputs that are *syntactically valid* and *semantically reasonable*, as in SD-VAE. This awaits future work.

## 4.3 Graph GANs

For *graph data* (see 3.5 Graph VAEs) we can denote a given *graph* as $G = (V; E)$, where $V = \{v_1, …, v_V\}$ represents the set of *vertices* (or *nodes*) and $E = \{e_{ij}\}_{i,j=1}^{V}$ represents the set of *edges*. For a given vertex $v_c$, we define $N(v_c)$ as the set of vertices directly connected to $v_c$. We further denote the underlying connectivity distribution for vertex $v_c$ as conditional probability $p(v | v_c)$. $N(v_c)$ can then be seen as a set of observed samples drawn from $p(v | v_c)$.

The *GraphGAN* [29] is a graph representation learning framework using the *node embeddings* approach. In GraphGAN, *adversarial learning* is based on the following scenario in which the generator network competes against an adversary:

- The *generator network* $g(v | v_c; \theta^{(g)})$ tries to approximate the underlying connectivity distribution $p(v | v_c)$ and generates the most likely vertices to be connected with $v_c$ from the vertex set V. (A *graph softmax* is used to overcome the limitations of the traditional softmax.)
- The *discriminator network* $D(v, v_c; \theta^{(d)})$ aims to discriminate the connectivity for the vertex pair $(v, v_c)$. It outputs a single scalar representing the probability of an edge existing between v and $v_c$.

Using node embedding approaches for generating entire graphs, however, can produce samples that don't preserve any of the patterns inherent to real graphs.

The *NetGAN* [30] architecture has the associated goals of learning from a single graph, generating discrete outputs, and staying invariant under node reordering. It generates graphs via *random walks*. NetGAN consists of two main components - a generator network and a discriminator network. The *generator network* is defined as a stochastic neural network with discrete outputs, whose goal is to generate synthetic random walks that are plausible in the input graph. At the same time, the *discriminator network* learns to distinguish the synthetic random walks from the real ones that come from the input graph. NetGAN preserves important *topological properties*, without having to explicitly specifying them in the model definition. Moreover, it can be used for generating new graphs with continuously varying characteristics using *latent space interpolation*.

## 5 Summary and Conclusion

Generative concept representations are probabilistic and generative in nature and can represent uncertainty, support integration of learning and reasoning, and are good for unsupervised and semi-supervised learning. They adopt the paradigm, and are the outcome, of probabilistic and generative deep learning.

Probabilistic and generative deep learning is based on probabilistic machine learning and probabilistic graphical models, while adopting the deep learning approach. Its core is deep generative models which utilizes latent variables and differentiable generator networks.

Variational autoencoders and generative adversarial networks can be used for learning generative concept representations. Both types of architectures have been enhanced or extended to handle data that are sequences, structured data or graphs.

Generative concept representations can be directly manipulated to generate new concepts with desired attributes. They are the foundation of creative applications.